\pdfoutput=1
\PassOptionsToPackage{dvipsnames}{xcolor}
\documentclass[11pt]{article}

\usepackage[final]{acl}

\usepackage{times}
\usepackage{latexsym}
\usepackage{amsmath}
\usepackage[T1]{fontenc}

\usepackage[utf8]{inputenc}

\usepackage{microtype}

\usepackage{inconsolata}

\usepackage{graphicx}
\usepackage{color}
\usepackage{xcolor}
\usepackage{xspace}
\usepackage{booktabs} 
\usepackage{multirow}
\usepackage{tabularx}
\usepackage{graphicx}
\usepackage{float}
\usepackage{makecell}
\usepackage{CJK}
\usepackage{changes}
\usepackage{varwidth}
\usepackage{soul, color, xcolor}
\usepackage{amsmath,bm}

\usepackage{changes}
\usepackage{subcaption}
\title{In-Context Former: Lightning-fast Compressing Context for Large Language Model}

\newcommand*{\affaddr}[1]{#1}

\newcommand*{\email}[1]{\texttt{#1}}

\author{Xiangfeng Wang, Zaiyi Chen, Tong Xu\thanks{Corresponding author.}, Zheyong Xie, Yongyi He, Enhong Chen\\
\affaddr{University of Science and Technology of China} \\
\email{\{xf9462, czy6516, xiezheyong, vagabond\}@mail.ustc.edu.cn, } \\
\email{\{tongxu, cheneh\}@ustc.edu.cn} \\
\url{https://github.com/wonderful9462/IC-Former}\\
}

\begin{document}

\newcommand{\ours}{IC-Former\xspace}

\maketitle

\begin{abstract}

With the rising popularity of Transformer-based large language models (LLMs), reducing their high inference costs has become a significant research focus. 
One effective approach is to compress the long input contexts. Existing methods typically leverage the self-attention mechanism of the LLM itself for context compression. 
While these methods have achieved notable results, the compression process still involves quadratic time complexity, which limits their applicability.
To mitigate this limitation, we propose the In-Context Former (IC-Former). 
Unlike previous methods, IC-Former does not depend on the target LLMs. Instead, it leverages the cross-attention mechanism and a small number of learnable digest tokens to directly condense information from the contextual word embeddings.
This approach significantly reduces inference time, which achieves linear growth in time complexity within the compression range.
Experimental results indicate that our method requires only 1/32 of the floating-point operations of the baseline during compression and improves processing speed by 68 to 112 times while achieving over 90\% of the baseline performance on evaluation metrics. Overall, our model effectively reduces compression costs and makes real-time compression scenarios feasible.

\end{abstract}

\section{Introduction}

In recent years, transformer-based ~\cite{attention} language models especially large language models (LLMs) have made significant strides in the field of natural language processing, demonstrating exceptional performance across a wide range of tasks. 
However, the self-attention mechanism in LLMs leads to high inference costs.
Previous work ~\cite{sparse,longformer,rmt,linear,memformer,longnet,xltransformer,rethinking,retro} has explored various approaches to reduce computational complexity by improving the self-attention mechanism of language models.
Although these strategies mitigate the overhead of long context processing, they inevitably introduce modifications to the original structure of LLMs, potentially impacting the capabilities of the original model \cite{lost}.

\begin{figure}
  \centering
  \includegraphics[width=0.9\linewidth]{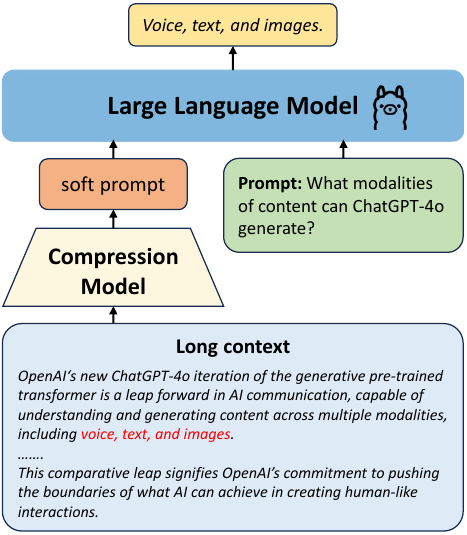}
  \vspace{0.em}
  \caption{Compressing long contexts into short soft prompts (vectors in embedding space) to improve inference efficiency.}
  \label{fig:intro}
  \vspace{-1.em}
\end{figure}

To better avoid modifications to the LLM structure, a more intuitive approach is to introduce a preliminary context compression process. These methods are based on a core assumption: most natural language texts contain redundant information, which makes context compression feasible. 
In early exploration, \citet{gist} have attempted to compress the instructions into short soft prompts. This method offers a novel perspective but still has limitations in long context compression. Later works \cite{autocompressor, icae} aim to further extend compression abilities for document-level long contexts, and achieved considerable results. 
As illustrated in Figure \ref{fig:intro}, these methods design compression models to condense lengthy contexts into short, context-rich soft prompts, which then serve as substitutes for the original context when input into the LLM.
However, these methods still suffer the issue of expensive time costs during the compression process. 
This limitation restricts their application in real-time compression scenarios, such as compressing retrieved ~\cite{realm, rpg} or real-time Internet documents ~\cite{asai2023self} immediately.

By reviewing previous works on compressors, we find that existing methods typically utilize the LLM as the encoder. While these methods fully utilize the powerful semantic understanding capabilities of LLM, they also suffer from rapidly increasing quadratic time complexity as the context lengthens. So is there a way to significantly reduce the theoretical complexity of compressors, with an acceptable decrease in performance?

Driven by this motivation, we design an efficient context compression model, the In-Context Former (IC-Former), which aims at optimizing resource consumption during the compression of long context in existing models.
This model is based on two assumptions regarding semantic content compression:
(1) Word embeddings already contain sufficient semantic information \cite{word2vec,cluster}, suggesting that interactions between embeddings may not be necessary prior to the extraction process.
(2) Learnable tokens within an elaborate structure can effectively aggregate information to a certain extent \cite{autocompressor, icae}.
Based on these assumptions, we try to discard the costly self-attention interaction of text content in previous models. Instead, we leverage the efficiency of the cross-attention mechanism for information extraction. This innovative strategy ensures that the computational overhead of compression grows linearly with the context length within the compression range, significantly enhancing compression efficiency compared to the previous methods. 

Specifically, our IC-Former consists of a few cross-attention blocks and some learnable digest tokens. Through this structure, the IC-Former leverages the digest tokens to extract information from lengthy contextual content and refine it into compact digest vectors. Subsequently, these digest vectors directly replace the original, verbose context and serve as input to LLMs while ensuring that the generated texts are faithful to the original context.

In the training phase, to effectively compress context, we follow the previous training paradigm  \cite{icae}, employing a strategy that combines pre-training and fine-tuning to optimize the IC-Former. 
During the pre-training phase, the IC-Former engages in a context reconstruction task. It generates digest vectors from which an LLM can reconstruct the original context. In the fine-tuning phase, we train the IC-Former on instruction data to ensure the generated digest vectors correctly respond to various context-related prompts.

Additionally, through theoretical calculations, we demonstrate that at a compression ratio of 4x, our IC-Former achieves only 1/32 of the floating-point operations required by the baseline. Experimental results further show that our method achieves a compression speed that is 68 to 112 times faster than the baseline while maintaining over 90\% of the baseline performance on evaluation metrics. This indicates a higher cost-effectiveness.

Overall, our contributions can be summarized in the following three points:
\begin{itemize}
    \item We propose the In-Context Former (IC-Former), a novel context compression model that can compress context to a quarter of its original length as a soft prompt while preserving most of original contextual information.
    
    \item The IC-Former is lightweight and efficient, with a parameter size that is 9\% of the target LLM. It achieves compression speeds 68 to 112 times faster than the baseline while maintaining more than 90\% of the baseline performance.
    
    \item We analyze the interaction between the IC-Former and the context, enhancing the interpretability of the IC-Former's compression process.
\end{itemize}

\section{Related Work}

\begin{figure*}[!t]
  \centering
  \includegraphics[width=\linewidth]{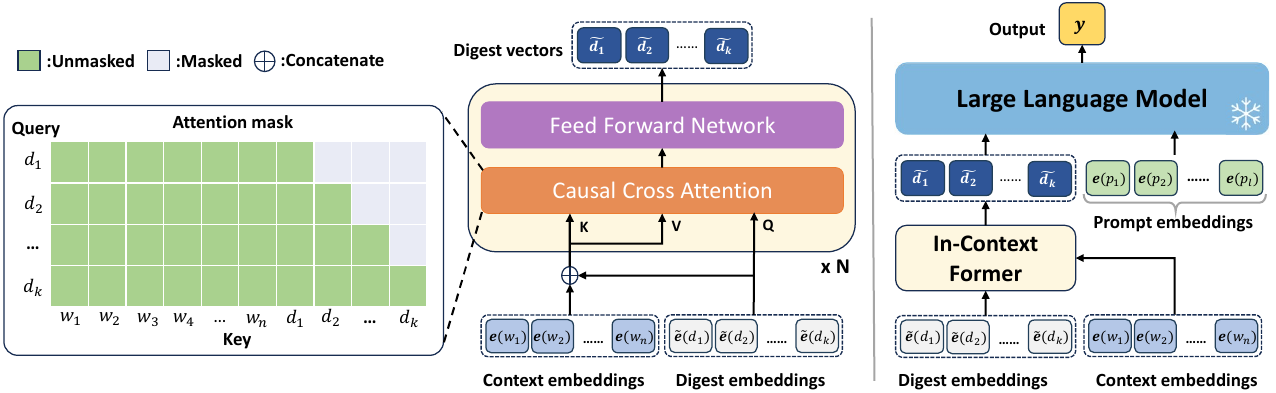}
  \vspace{-1.em}
  \caption{\textbf{Left:} Model architecture of In-Context Former. In-Context Former utilizes a set of learnable digest embeddings to condense the information of context and generates digest vectors. And we apply causal attention masks for digest tokens. \textbf{Right:} Overview of In-Context Former's framework.}
  \label{fig:model}
  \vspace{-1.em}
\end{figure*}

\noindent\textbf{Soft prompt compression}
\citet{prompt_comp} propose to learn a compact soft prompt ~\cite{prompt_tuning} to represent the original natural language prompt. 
They align the model predictions that are based on the original prompt and those conditioned on the soft prompt by optimizing KL divergence ~\cite{kl}.
As a result, \citet{prompt_comp} discover that the trained soft prompt retain high-level semantic information and can be utilized to control generation. 
However, this approach suffers high computational costs as it requires retraining a new soft prompt for each new context. 
In contrast, our method can predict the soft prompt corresponding to the input context.

\noindent\textbf{Context distillation}
Another related work \cite{snell, ask} focuses on distilling the contextual information such as instruction into a student model without prompting. 
\citet{gist} propose GIST to compress prompts into gist tokens, which can be viewed as key-value attention prefixes. 
Nonetheless, this approach did not address the long context issue as it is limited to compressing short prompts. 
In addition, this method requires updating the parameters of language model, which differs from our method. 
Our method keeps the language model fixed and therefore preserves its capability.

\noindent\textbf{Context compression}
\citet{autocompressor} propose AutoCompressors to compress long text into summary vectors recursively. 
However, the compression procedure is sophisticated and LLMs are still required to be fine-tuned to generate summary vectors. 
ICAE \cite{icae} is the most closely related study to our research. 
ICAE compresses context into short memory slots, with a small number of additional parameters by the LoRA \cite{lora} approach with a fixed LLM. 
However, both AutoCompressors and ICAE employ self-attention to integrate contextual information, resulting in a quadratic complexity with respect to the length of context. 
Instead, our model does not incorporate contextual interactions and reduces both time and space complexities, striking a balance between efficiency and performance.

\section{Method}

\subsection{Task Formulation} \label{sec:task formulation}
Context compression aims to transform lengthy contexts into brief, compact representations while endeavoring to preserve the fundamental semantics and integrity of the original contexts. 

Formally, we define the original context that is to be compressed as $w=(w_1, w_2,..., w_n)$, where $w_i$ represents the $i$-th token of context and $n$ is the number of tokens in context. 
Then, we denote $\boldsymbol{e}(\cdot)$ as the word embedding lookup in the LLM and $\tilde{\boldsymbol{e}}(\cdot)$ as the learnable embeddings of soft tokens. 
A context compressor model $\Theta$ utilizes the embeddings of soft tokens $\widetilde{\boldsymbol{e}}(d)=(\tilde{\boldsymbol{e}}(d_1), \tilde{\boldsymbol{e}}(d_2), ..., \tilde{\boldsymbol{e}}(d_k))$ and context embeddings $\boldsymbol{e}(w)=(\boldsymbol{e}(w_1), \boldsymbol{e}(w_2), ..., \boldsymbol{e}(w_n))$ to generate compact representations $\widetilde{\boldsymbol{d}}=(\widetilde{\boldsymbol{d}_1}, \widetilde{\boldsymbol{d}_2}, ..., \widetilde{\boldsymbol{d}_k})$ of context, where $k$ is the length of compressed context and $k\ll n$. 

The condensed vectors $\widetilde{\boldsymbol{d}}$ can substitute the original context and be combined with other prompt $\boldsymbol{e}(p)=(\boldsymbol{e}(p_1),..., \boldsymbol{e}(p_l))$ for input to an LLM $\Phi$.
The output $y=(y_1,...,y_m)$ remains faithful to the content of the original context $w$.

\begin{figure*}[!ht]
  \centering
  \includegraphics[width=0.85\linewidth]{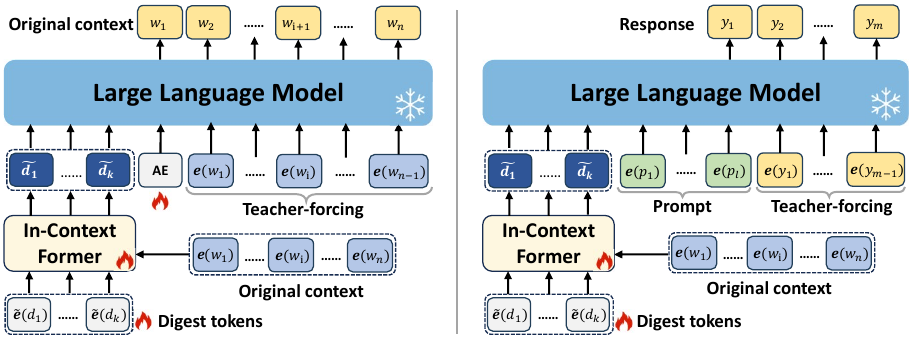}
  \vspace{0.em}
  \caption{\textbf{Left:} Pretraining stage. IC-Former learns to generate digest vectors such that, when these vectors and a special token AE are jointly fed into an LLM, the LLM reproduces the original context. \textbf{Right:} Instruction fine-tuning stage. Training IC-Former to generate digest vectors capable of correctly responding to prompts.}
  \label{fig:training}
  \vspace{-1.em}
\end{figure*}

\subsection{In-Context Former} \label{sec:ic-former}

As illustrated in Figure \ref{fig:model}, IC-Former consists of a few cross-attention layers and a set of learnable soft tokens, which are named digest tokens. The IC-Former utilizes context tokens and digest tokens as inputs, leveraging a causal cross-attention mechanism to condense the context information into digest vectors. 
Subsequent sections will detail the attention computation process, attention masks, and positional embeddings.

\noindent\textbf{Attention computation}
When compressing a long context, the context tokens are concatenated with digest tokens and subsequently mapped into embeddings, which serve as key and value in the cross-attention layer. 
Meanwhile, the embeddings of digest tokens serve as query to interact with both context embeddings and digest embeddings. 
To be specific, the $Q$, $K$ and $V$ in IC-Former can be computed as:
\begin{align}
    \label{Q} Q&=W_{Q}\ \tilde{\boldsymbol{e}}(d)^{T} \\
    \label{K} K&=W_{K} \left[\boldsymbol{e}(w); \tilde{\boldsymbol{e}}(d)\right]^{T} \\
    \label{V} V&=W_{V} \left[\boldsymbol{e}(w); \tilde{\boldsymbol{e}}(d)\right]^{T} 
\end{align}

Then we employ the cross-attention mechanism to condense contextual information, as this approach has been empirically validated effective in multimodal information extraction.~\cite{blip2,mplugowl,minigpt,qwen}.

\noindent\textbf{Attention masks} 
As depicted in Figure \ref{fig:model}, our design for attention masks allows digest tokens to attend to all context tokens as well as preceding digest tokens, thereby mitigating the deficiency of interaction among context tokens.

Additionally, it can be observed from the attention matrix that given a context length of $n$ and a target compression length of $k$, the time complexity and space complexity of our method are both $\mathcal{O}(kn + k^2)\sim \mathcal{O}(kn)$.
This indicates that the complexity of this model grows linearly with the increase of context.

\noindent\textbf{Positional embeddings} We recognize that the pure cross-attention mechanism does not capture the relative positional relationships among tokens within the context. 
This implies swapping any two tokens in the context results in an identical digest vector, which does not align with our expectations.
To address this, we applied RoPE \cite{rope} to represent the relative positional relations within the context tokens.

We denote the positional embeddings of the $n$th token in the sequence as $\mathrm{RoPE}(n)$ and is abbreviated as $\mathrm{R}_n$.
\begin{align}\label{RoPE}
&\mathrm{RoPE}(n)=\left[ 
\begin{matrix} 
{R_{n}^{(0)}} & {} & {} \\ 
{} & {R_{n}^{(1)}} & {} \\ 
{} & {} & {\ddots} & {} \\ 
{} & {} & {} & {R_{n}^{(\frac{h}{2}-1)}} \\ 
\end{matrix} \right], \notag\\
&\mathrm{where}\ R_{n}^{(i)}=\left[ 
\begin{matrix} 
{\operatorname{cos}(n \theta^{i})} & {-\operatorname{sin} (n\theta^{i})} \\ 
{\operatorname{sin} (n\theta^{i})} & {\operatorname{cos} (n\theta^{i})} \\ 
\end{matrix} \right]
\end{align}
In the Equation.\ref{RoPE}, $\theta=\theta_{base}^{-\frac{2}{h}}$ where $\theta_{base}$ is a hyper-parameter and $h$ is the hidden size and assumed to be even. 
We restate Equation.\ref{Q} \& \ref{K} as follows:
\begin{align}
    Q&=(\boldsymbol{q}_1, \boldsymbol{q}_2,...,\boldsymbol{q}_k)\\ 
    K&=(\boldsymbol{k}_1, ..., \boldsymbol{k}_n, \boldsymbol{k}_{n+1},..., \boldsymbol{k}_{n+k})
\end{align}
We allocate positional embeddings as if placing the digest tokens subsequent to the context tokens as demonstrated in Equation.\ref{RoPE Q} \& \ref{RoPE K}.
\begin{align}\label{RoPE Q}
    Q_{\mathrm{RoPE}}&=(\mathrm{R}_{n+1}\boldsymbol{q}_1, \mathrm{R}_{n+2}\boldsymbol{q}_2,...,\mathrm{R}_{n+k}\boldsymbol{q}_k) \\
    K_{\mathrm{RoPE}}&=(\mathrm{R}_{1}\boldsymbol{k}_1, ... , \mathrm{R}_{n}\boldsymbol{k}_n,...,\mathrm{R}_{n+k}\boldsymbol{k}_{n+k})\label{RoPE K}
\end{align}
The RoPE manifests the relative positional relationships through the inner product between $Q_{\mathrm{RoPE}}$ and $K_{\mathrm{RoPE}}$:
\begin{equation}
    (\mathrm{R}_i\boldsymbol{q})^{T}(\mathrm{R}_j\boldsymbol{k})=\boldsymbol{q}^{T}\mathrm{R}_i^{T}\mathrm{R}_j\boldsymbol{k}=\boldsymbol{q}^{T}\mathrm{R}_{j-i}\boldsymbol{k}
\end{equation}
In this manner, each digest token is capable of perceiving the relative positions of both context tokens and other digest tokens.

\subsection{Training process} \label{sec:training}
This section introduces the training objectives of IC-Former, including pretraining and instruction fine-tuning, and a divide-and-conquer training strategy when dealing with too long contexts.

\noindent\textbf{Pretraining}
Previous works \cite{ae, vae, vqvae, icae} have demonstrated that autoencoding tasks can benefit models to effectively condense and encode information.
We adopt this approach to pretrain our IC-Former by using a text reconstruction task.
The objective of this task is to leverage digest vectors, which are extracted from compressed contexts, to reconstruct the original contexts.
As illustrated in Figure \ref{fig:training}, the context tokens are compressed into digest vectors by IC-Former and then serve as input to LLM with a special token "[AE]" to indicate the autoencoding task.

To make LLM reconstruct the original context $w$ conditioned on the digest vectors $\widetilde{\boldsymbol{d}}$, we optimize IC-Former $\Theta$ and digest embeddings $\widetilde{\boldsymbol{e}}(d)$ by minimizing negative log-likelihood of context $w$.
The pretraining objective can be written as:
\begin{align}
    \mathcal{L}_{\mathrm{AE}}&=-\log p\left(w|\widetilde{\boldsymbol{d}_1},...,\widetilde{\boldsymbol{d}_k}; \Phi\right) \notag \\
    &=-\log p\left(w|d_1,...,d_k;\widetilde{\boldsymbol{e}}; \Theta; \Phi \right)
\end{align}
This reconstruction task forces IC-Former to focus on each token in context, thereby preserving all context information. 
The analysis on pretraining in Section \ref{sec:analysis} demonstrates that this task can help IC-Former learn to aggregate contextual information.

\noindent\textbf{Instruction fine-tuning}\quad
After the pretraining phase, IC-Former has effectively learned to meticulously attend to context. 
However, to ensure that the compressed digest vectors appropriately respond to various prompts, further instruction fine-tuning \cite{instruction} of IC-Former is necessary.
As shown in Figure \ref{fig:training}, we input the digest vectors generated from IC-Former along with the prompt embeddings into the LLM.
Similarly, by optimizing IC-Former $\Theta$ and digest embeddings $\widetilde{\boldsymbol{e}}(d)$, we minimize the negative log-likelihood of the expected output $y$:
\begin{align}
    \mathcal{L}_{\mathrm{FT}}&=-\log p(y|\widetilde{\boldsymbol{d}_1},...,\widetilde{\boldsymbol{d}_k};p_1,...,p_l;\Theta;\Phi) \notag \\
    &=-\log p(y|d_1,...,d_k;p_1,...,p_l;\widetilde{\boldsymbol{e}};\Theta;\Phi)
\end{align}

\begin{figure}[!t]
  \centering
  \includegraphics[width=0.9\linewidth]{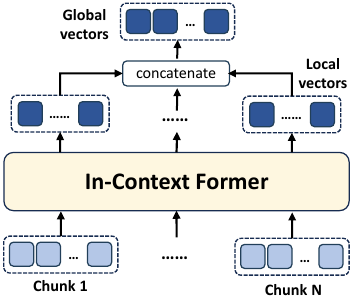}
  \vspace{0.em}
  \caption{The excessively long contexts are broken into chunks, which are then compressed and concatenated.}
  \label{fig:divide conquer}
  \vspace{-1.em}
\end{figure}

\noindent\textbf{Divide and conquer} 
When the context length exceeds the compression limit, a divide-and-conquer strategy ~\cite{unlimiformer, hierarchical,walking} proves to be effective.
We first uniformly split the context into several chunks of acceptable length. 
Each of these chunks is then compressed individually to obtain local vectors. 
As illustrated in Figure \ref{fig:divide conquer}, we subsequently concatenate all these local vectors to form the global vectors.
This strategy is applied in both the training and inference phases.

\section{Experiments}\label{sec:experiments}
\subsection{Experimental setting}
This section introduces the experimental setting including data, baseline, and model configuration.

\noindent\textbf{Data}
Due to resource constraints, we pretrain IC-Former using a subset of the Pile \cite{pile} dataset, comprising approximately 2.29 million text entries.
In the fine-tuning phase, we employed the PwC (Prompt-with-Context) dataset \cite{icae}, which includes contexts accompanied by corresponding questions. 
This dataset is suitable for evaluating the compressor's ability to preserve contextual information.
For each context, the dataset provides ten specific and five general questions. 
For evaluation convenience, we select the ten specific questions to evaluate as their answers are relatively more definitive. 
\begin{table*}[!ht]
\vspace{0.em}
\centering
\resizebox{\textwidth}{!}{
\begin{tabular}[width=\linewidth]{cccccc}
\toprule
\makecell{\textbf{Input} \\ \textbf{(Batchsize$\times$Length)}} & \textbf{Method} & \makecell{\textbf{Memory} \\ \textbf{(GB)}} & \makecell{\textbf{Compression}\\ \textbf{Time (s)}} & \makecell{\textbf{Inference}\\ \textbf{Time (s)}} & \makecell{\textbf{Total} \\ \textbf{Time (s)}} \\
 \midrule
\multirow{3}{*}{$8\times2048$} & LLM & 35.96 & - & 1.845 & 1.845 \\ 
& LLM+ICAE & 19.76 & 3.268 & 0.314 & 3.582 \\ 
& LLM+IC-Former & \textbf{15.96 / 2.38} & \textbf{0.029 (112$\times$)} & 0.314 & \textbf{0.343 (5.3$\times$)} \\ 
\midrule
\multirow{3}{*}{$8\times512$} & LLM & 17.46 & - & 0.318 & 0.318 \\ 
& LLM+ICAE & 19.76 & 0.476 & 0.079 & 0.555 \\ 
& LLM+IC-Former & \textbf{15.82 / 2.28} & \textbf{0.007 (68$\times$)} & 0.079 & \textbf{0.086 (3.7$\times$)} \\ 
 \midrule
\multirow{3}{*}{$32\times512$} & LLM & 29.07 & - & 1.186 & 1.186 \\ 
& LLM+ICAE & 38.74 & 1.848 & 0.288 & 2.136 \\  
& LLM+IC-Former & \textbf{18.98 / 3.52} & \textbf{0.017 (108$\times$)} & 0.289 & \textbf{0.306 (3.8$\times$)} \\ 
\bottomrule
\end{tabular}}
\caption{Compression and inference overhead. Inference time refers to the period required to perform a forward pass, utilizing either original context embeddings or compressed vectors as input to the LLM. Memory denotes the peak GPU memory usage during the compression and inference processes. Additionally, we quantify the memory utilization when employing IC-Former for compression independently (right of the /).}
\label{tab:overhead}
\vspace{-1.em}
\end{table*}

\noindent\textbf{Baseline}
We select ICAE as our baseline for comparison, because the motivations behind other related works are distinct from ours. 
For instance, AutoCompressors fine-tune LLMs and focus on stability in long-context modeling rather than on restoring details in compressed text. 
Likewise, GIST also modifies model parameters, and its strength lies in compressing instruction information rather than long context. We replicate ICAE on this dataset.

\noindent\textbf{Model configuration}
We use Llama2-7b-chat \cite{llama2} as the target LLM for evaluation. Both attention and feed-forward network modules of IC-Former have the same hidden size as Llama2-7b-chat.
The default number of digest tokens $k$ is set to 128 unless otherwise specified.
Furthermore, IC-Former consists of only three transformer layers and includes approximately 607M parameters, encompassing the digest embeddings.

\begin{table}[!t]
\centering
\resizebox{\linewidth}{!}{
\begin{tabular}{ccc}
\toprule
\textbf{Method} & \makecell{\textbf{Time\&Space} \\ \textbf{Complexity}} & \makecell{\textbf{Theoretical} \\ \textbf{FLOPs}}\\
\midrule
ICAE & $\mathcal{O}(n^2+2kn)$ & $8.50\times 10^{12}$\\ 
IC-Former & $\mathcal{O}(kn)$ & $2.62\times 10^{11}(\sim \frac{1}{32})$\\
\bottomrule
\end{tabular}}
\caption{Complexity analysis. The theoretical FLOPs represent the computational cost incurred when compressing a context of length 512 into 128 vectors for the Llama2-7b-chat model. For further details, see the Appendix \ref{app:analysis}.}
\label{tab:complexity}
\vspace{-1.em}
\end{table}

\subsection{Experiment Results}
\subsubsection{Compression \& Inference Efficiency}
Firstly we analyze the theoretical time-space complexity of the IC-Former and baseline method and the floating point operations (FLOPs) required to compress 512 tokens to a length of 128.
As illustrated in Table \ref{tab:complexity}, our approach significantly reduces both the temporal and spatial overhead compared to the baseline. 
In experiments involving compression of contexts with a length of 512, the required FLOPs are merely 1/32 of those needed by the baseline method.

We further assess and compare the compression time and memory utilization of IC-Former during actual compression processes with the baseline model. Experimental results indicate that our IC-Former significantly outperforms existing methods in terms of both temporal efficiency and spatial occupancy.

As shown in Table \ref{tab:overhead}, our IC-Former has the lowest memory usage during compression among the compared models. 
Additionally, IC-Former's compression process does not depend on the target LLM, enabling it to perform compression independently and achieve over 88\% memory savings relative to the baseline.
In terms of compression time, our method is 68 to 112 times faster than the baseline, rendering the compression overhead negligible compared to the inference time of the target LLM.
In scenarios where compression is followed by inference, our method achieves approximately four times faster processing than directly inferring using the original context, whereas the baseline method consumes even more time. 
Our approach thus offers a viable solution for real-time compression scenarios.
\begin{table}[!t]
\centering
\belowrulesep=0pt\aboverulesep=0pt
\captionsetup{skip=10pt} 
\setlength{\belowcaptionskip}{-10pt}
\resizebox{0.48\textwidth}{!}{
\begin{tabular}{c|cc|cc}
\toprule
\specialrule{0em}{1pt}{0pt}
\multirow{2}{*}{\textbf{Length}} & \multicolumn{2}{c|}{\textbf{BLEU-4}} & \multicolumn{2}{c}{\textbf{Loss}} \\
\cmidrule(lr){2-3}  \cmidrule(lr){4-5} 
& ICAE & IC-Former & ICAE & IC-Former \\ 
\specialrule{0em}{1pt}{0pt}
\midrule
\specialrule{0em}{0pt}{1pt}
100 & \textbf{0.9967} & 0.9965 & \textbf{0.1461} & 0.1789 \\
200 & 0.9969 & \textbf{0.9972} & 0.0971 & \textbf{0.0851} \\
300 & \textbf{0.9974} & 0.9971 & 0.0602 & \textbf{0.0558} \\
400 & 0.9889 & \textbf{0.9892} & 0.0499 & \textbf{0.0483} \\
500 & 0.9654 & \textbf{0.9689} & 0.1116 & \textbf{0.1078} \\
\specialrule{0em}{1pt}{1pt}
\bottomrule
\end{tabular}}
\caption{Results of BLEU-4 scores and cross-entropy loss between reconstructed context and original context across different context lengths.}
\label{tab:pretrain}
\vspace{-1.em}
\end{table}
\begin{table*}[t]
\centering
\resizebox{\textwidth}{!}{
\begin{tabular}{ccclcccccc}
\toprule
\multirow{2.4}{*}{\textbf{Input content}} & \multicolumn{3}{c}{ \textbf{ROUGE-1} } & \multicolumn{3}{c}{ \textbf{ROUGE-2} } & \multicolumn{3}{c}{ \textbf{ROUGE-L} } \\
\cmidrule(lr){2-4} \cmidrule(lr){5-7} \cmidrule(lr){8-10}
& P & R & F1 & P & R & F1 & P & R & F1 \\ 
\midrule

512 original context tokens & 0.456 & \textbf{0.635} & 0.501 & 0.300 & \textbf{0.438} & 0.331 & 0.426 & \textbf{0.594} & 0.468 \\
128 memory slots (ICAE) & \textbf{0.592} & 0.561 & \textbf{0.555} & \textbf{0.404} & 0.385 & \textbf{0.377} & \textbf{0.553} & 0.525 & \textbf{0.519} \\
128 digest vectors (IC-Former) & 0.554 & 0.520 & 0.516 & 0.374 & 0.355 & 0.348 & 0.517 & 0.487 & 0.482 \\
(performance ratio)& 93.6\% & 92.7\% & 93.0\% & 92.6\% & 92.2\% & 92.3\% & 93.5\% & 92.8\% & 92.9\% \\
\midrule
64 digest vectors & 0.384 & 0.412 & 0.377 & 0.211 & 0.234 & 0.209 & 0.349 & 0.375 & 0.343 \\
64+64 digest vectors & 0.545 & 0.498 & 0.500 & 0.358 & 0.330 & 0.327 & 0.507 & 0.464 & 0.465 \\
128 digest vectors & \textbf{0.554} & \textbf{0.520} & \textbf{0.516} & \textbf{0.374} & \textbf{0.355} & \textbf{0.348} & \textbf{0.517} & \textbf{0.487} & \textbf{0.482} \\
\midrule
128 digest vectors (w/o pretrain) & 0.431 & 0.381 & 0.389 & 0.234 & 0.211 & 0.212 & 0.393 & 0.349 & 0.355 \\
\bottomrule
\end{tabular}}
\caption{Evaluation results on PwC test set. The first four rows of the table compare the performance of our method with other baseline models, and the performance ratio means the ratio of our IC-Former to the ICAE. The second three rows demonstrate the performance variations when different compression strategies are implemented, where "64+64" represents a divide-and-conquer approach. The last row reveals the impact of \textbf{ablation} pre-training on performance.}
\label{tab:finetune}
\vspace{-1.em}
\end{table*}

\subsubsection{Pretraining: Context Reconstruction}
\begin{table}[!t]
\centering
\begin{tabular}{c|c|c}
\bottomrule
\textbf{Text type} & \textbf{BLEU} & \textbf{Loss}  \\
\hline
Normal text & 0.9006 & 0.125 \\ 
Reversed text & 0.6652 & 1.803 \\
Patterned random text & 0.1347 & 4.401 \\
Completely random text & 0.0080 & 8.137 \\
\toprule
\end{tabular}
\caption{Reconstruction results for texts with varying degrees of randomness, with randomness increasing from top to bottom. The patterned text is generated by adding 1 to each token\_id of normal text. All texts above are compressed from length of 512 to 128.}
\label{tab:random}
\vspace{-1.em}
\end{table}
We evaluate the pretraining performance of IC-Former, focusing on its ability to reconstruct the original context.
To measure the discrepancies between the reconstructed text and the original, we utilize BLEU \cite{bleu} and cross-entropy loss as metrics.

As shown in Table \ref{tab:pretrain}, the reconstructed context by IC-Former exhibits minimal discrepancies when compared to the original context. 
For a context length of less than 400, the BLEU-4 score reaches 0.99, and the cross-entropy loss hovers around 0.05. 
When the context length is extended to 500, the BLEU score maintains a high value of 0.96, and the cross-entropy loss is approximately 0.1. 
These results suggest that IC-Former effectively captures the contextual information, achieving a 4x compression ratio while maintaining performance comparable to the baseline.

\begin{figure}[!t]
  \centering
  \includegraphics[width=1\linewidth]{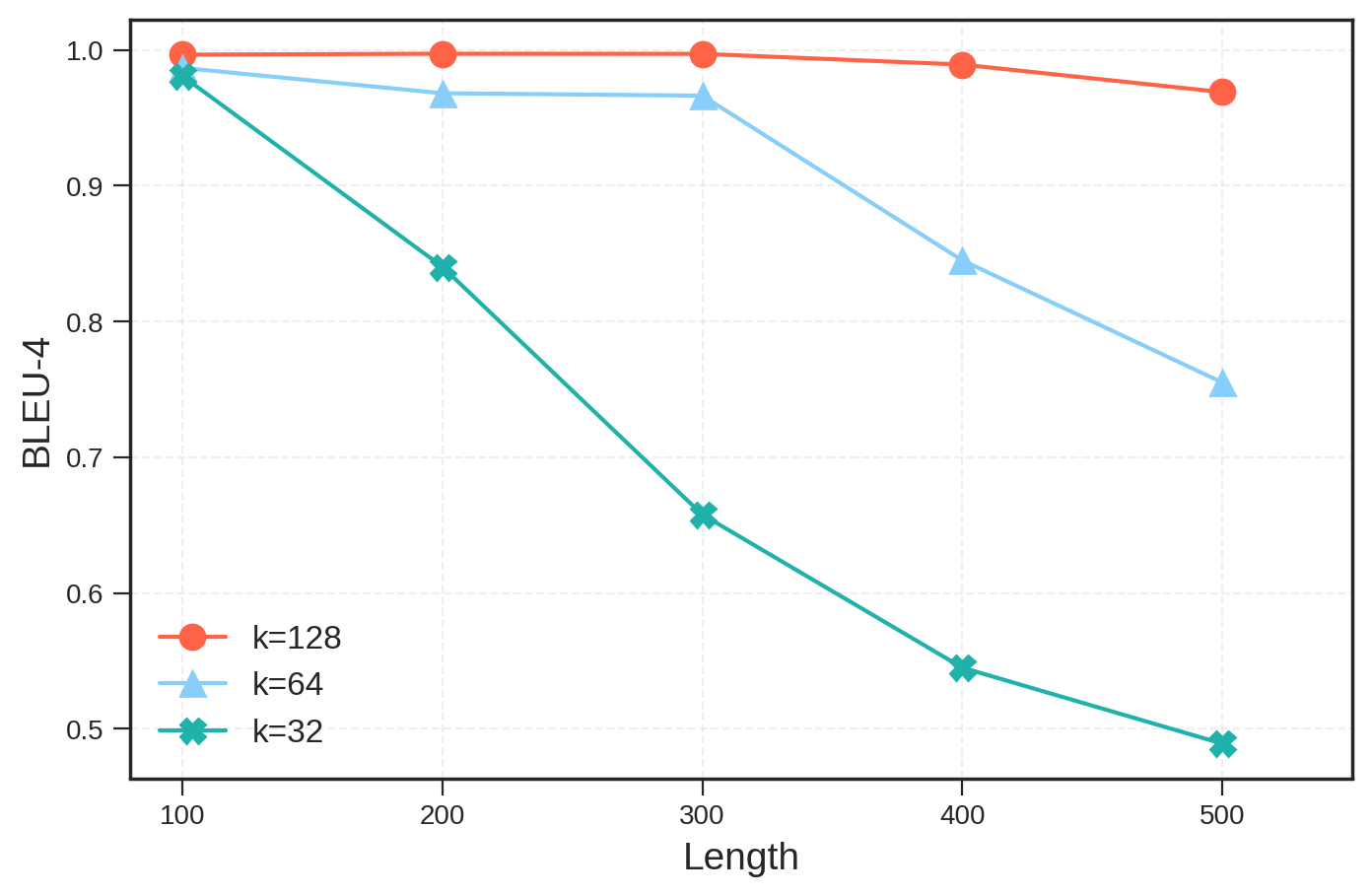}
  \vspace{-1.em}
  \caption{BLEU-4 for different digest token lengths $k$.}
  \label{fig:length_bleu}
  \vspace{-1.em}
\end{figure}

Then we explore the impact of digest tokens length $k$ on the reconstruction task.
As shown in Figure \ref{fig:length_bleu}, it is not surprising that the quality of the reconstructed text deteriorates as $k$ decreases.

Additionally, we attempt to use IC-Former to compress texts with various levels of randomness and analyze the reconstruction results.
As observed from Table \ref{tab:random}, the reconstruction performance of IC-Former progressively declines as the randomness of the text increases.
This phenomenon may suggest that IC-Former primarily achieves information compression through semantic understanding rather than mere rote memorization. Further analysis is conducted in Section \ref{sec:analysis}.

\subsubsection{Performance on Downstream Task}
In this section, we evaluate the model's performance on the PwC dataset. Although our model can achieve good results based on the BLEU metric, considering that BLEU is more susceptible to response length, we ultimately choose the ROUGE metric~\cite{rouge} to evaluate the performance of our model, which more faithfully reflects the original content of the text.
We compare the performance of various context compression models by keeping the target LLM frozen and substituting the context with different vectors.

As illustrated in the first row of Table \ref{tab:finetune}, our method achieves over 92\% of the baseline performance while significantly surpassing the baseline model in terms of compression speed.
The second row of the table compares the performance of digest vectors of varying lengths, including the compression of 512 context tokens into 64 digest vectors and their subsequent division and compression into two sets of 64 digest vectors each, as discussed in Section \ref{sec:training} under the strategy of divide-and-conquer. 
It can be observed that compared to directly compressing 512 context tokens into 128 digest vectors, the approach of divide-and-conquer results in a slight performance degradation. 
However, this performance loss is acceptable when compared to the costs associated with retraining a model to accommodate longer digest embeddings.
Additionally, we utilize an ablation study to demonstrate the efficacy of pretraining. 
IC-Former without pretraining performs poorly in capturing contextual information and is more prone to generating hallucinations. (See examples in Appendix \ref{app:case}).

\begin{figure}[!t]
  \centering
  \includegraphics[width=1\linewidth]{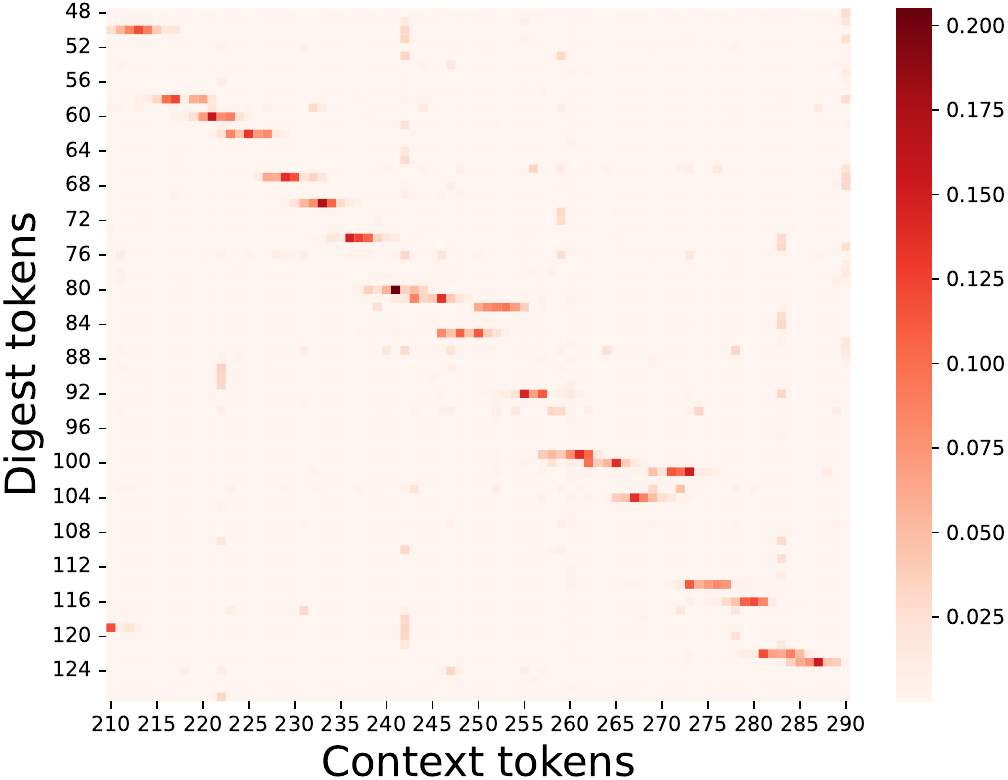}
  \vspace{-0.em}
  \caption{A part of attention map in the last layer of IC-Former. The horizontal axis represents context tokens acting as key and the vertical axis represents digest tokens acting as query. For complete attention map, see Appendix \ref{app:attention map}.}
  \label{fig:local_attention}
  \vspace{-1.em}
\end{figure}

\subsection{Analysis}\label{sec:analysis}
To better understand the working principles of IC-Former, we conducted further visualization analysis based on the attention map.

\noindent\textbf{Neighbourhood information aggregation}
We average the attention scores of all attention heads in the third layer (last layer) of the IC-Former to obtain an attention map.
It can be observed from Figure \ref{fig:local_attention} that each digest token attends to 3 to 5 consecutive context tokens, and digest tokens focus on the context tokens in accordance with their sequential order, which presents a backslash shape pattern.
It is worth mentioning that the non-pretrained IC-Former does not exhibit this phenomenon (See examples in Appendix \ref{app:attention map}).
These phenomena indicate that IC-Former compresses context by aggregating information from adjacent tokens and integrating it into digest vectors.
Moreover, the application of positional embeddings ensures that digest tokens attend to context in a sequential manner.

\begin{table}[!t]
\centering
\small
\begin{tabularx}{\columnwidth}{|>{\raggedright\arraybackslash}X|}
\hline
\textbf{An example of context} \\ 
\hline
\color{red}A \color{blue}large language model (LL\color{red}M) \color{Green}is a \color{blue}computational model notable \color{Green}for \color{red}its ability \color{Green}to \color{red}achieve \color{blue}general\color{gray}-\color{blue}purpose language \color{red}generation \color{Green}and other \color{red}natural language \color{blue}processing tasks such \color{red}as \color{blue}classification\color{Green}. \color{red}Based \color{Green}on \color{blue}language models\color{Green}, \color{red}L\color{blue}LM\color{gray}s \color{red}acquire \color{Green}these \color{blue}abilities \color{gray}by \color{blue}learning statistical \color{red}relationships \color{gray}from \color{blue}vast amounts \color{Green}of \color{blue}text \color{red}during \color{blue}a computation\color{red}ally int\color{blue}ensive self-supervised \color{Green}and \color{blue}semi\color{gray}-\color{blue}supervised training process\color{Green}. \color{blue}LL\color{red}M\color{Green}s \color{blue}can \color{Green}be \color{blue}used \color{red}for \color{blue}text generation\color{Green}, a \color{blue}form \color{Green}of gener\color{gray}ative \color{red}AI\color{Green}, \color{red}by \color{blue}taking \color{red}an \color{blue}input text \color{Green}and \color{blue}repeatedly pGreenict\color{Green}ing the \color{blue}next token or word\color{Green}. \color{blue}LLM\color{Green}s are \color{blue}artificial neural networks \color{gray}that \color{blue}utilize \color{Green}the \color{blue}transformer architecture\color{Green}, \color{blue}invented \color{Green}in \color{blue}20\color{Green}17. The \color{blue}largest \color{Green}and most \color{blue}capable \color{red}L\color{blue}LMs\color{Green}, \color{blue}as \color{Green}of \color{blue}June 2\color{Green}0\color{blue}24\color{Green}, are \color{blue}built \color{red}with \color{Green}a \color{blue}decoder\color{gray}-\color{blue}only transformer\color{gray}-\color{blue}based architecture\color{Green}, \color{blue}which enables efficient processing \color{Green}and \color{blue}generation \color{Green}of \color{blue}large-scale text data\color{Green}. \color{blue}Larger models such \color{gray}as \color{red}G\color{blue}PT-\color{Green}3 \color{blue}have demonstrated \color{Green}the \color{blue}ability \color{Green}to \color{blue}achieve similar results through prompt engineering\color{Green}, \color{blue}which involves craft\color{gray}ing \color{blue}specific input prompts \color{Green}to \color{blue}guide \color{Green}the \color{blue}model's responses\color{Green}. \\ 
\hline
\end{tabularx}
\caption{The context tokens that are most attended to by digest tokens across layers. The color of each token is determined by the layer when it is \textbf{initially} attended to. \textcolor{Green}{Green}, \textcolor{blue}{blue}, and \textcolor{red}{red} denote the first, second and third layer respectively. \textcolor{gray}{Gray} indicates tokens that are never attended to.}
\label{tab:case}
\vspace{-1.em}
\end{table}

\noindent\textbf{Layer-wise semantic diversification}
Thanks to IC-Former being composed of merely three layers, we are able to conduct a detailed analysis of each layer. 
We examine each layer of the IC-Former to identify the top five context tokens with the highest attention scores for each digest token.

As illustrated in Table \ref{tab:case}, it can be observed that in the first layer, digest tokens mainly focus on prepositions, articles, be-verb, and punctuation marks. 
As we proceed to the second layer, digest tokens start to extend their focus to verbs, nouns, adjectives, and adverbs. 
The third layer continues this trend based on the second layer, further broadening the range of grammatical categories of tokens covered, encompassing a more extensive context.
This implies that IC-Former might rely on semantic structures to compress context effectively.

\section{Conclusion}
In this paper, we propose the In-Context Former (IC-Former), a novel context compression model, which can efficiently condense contextual information into digest vectors in a linear complexity by removing irrelevant interaction processing. 
Moreover, our proposed IC-Former utilizes the cross-attention mechanism to enhance the extraction ability of digest tokens.
Our experimental results demonstrate that IC-Former significantly reduces time and space complexity while preserving contextual semantics, thereby supporting broader applications requiring extensive context.

\section*{Limitations}
\begin{itemize}
    \item [1.] We only apply IC-Former to the Llama2-7b-chat model. Future efforts will involve conducting experiments on larger-scale models to explore further potential. It is anticipated that the increased hidden size in larger models will continue to enhance the performance of the IC-Former.
    \item [2.] 
    Although our method is capable of handling longer texts in implementation, we did not conduct compression experiments on longer contextual content to more comprehensively validate the method's performance due to resource constraints.
    \item [3.] Despite our model significantly outperforming the baseline in terms of efficiency, it has not surpassed the baseline's performance in downstream tasks. Our future work will aim to enhance performance in scenarios that are less sensitive to real-time requirements.
\end{itemize}

\section*{Acknowledgments}
This work was supported by the grants from National Natural Science Foundation of China (No.62222213, 62072423).

\bibliography{custom}

\clearpage
\appendix
\section{Experiment Details}\label{app:exp}
\subsection{Model Configuration}
We show the detailed configuration of our IC-Former model in Table \ref{tab:model config}.

\begin{table}[!h]
\centering
\begin{tabular}{l|c}
\toprule
\textbf{Hyperparameter} & \textbf{Value}\\
\midrule
theta base & 10000.0 \\ 
hidden size & 4096 \\ 
layer number & 3 \\
rms norm eps & 1e-6 \\
initializer range & 0.02 \\
activate function & silu \\
intermediate size & 11008 \\
digest tokens number & 128 \\
attention heads number & 32 \\
max position embeddings & 2048 \\
\bottomrule
\end{tabular}
\caption{Detailed configuration of IC-Former.}
\label{tab:model config}
\vspace{-1.em}
\end{table}

\subsection{Training Configuration}
We show the detailed configuration of pretraining and fine-tuning in Table \ref{tab:training config1} \& \ref{tab:training config2}.

\begin{table}[!ht]
\centering
\begin{tabular}{l|c}
\toprule
\textbf{Hyperparameter} & \textbf{Value}\\
\midrule
optimizer & AdamW \\ 
learning rate & 1e-4\\
batch size & 1 \\ 
gradient accumulation & 16\\
clip norm & 2.0 \\
training steps & 9.3k \\
dtype & bfloat16 \\
\bottomrule
\end{tabular}
\caption{Detailed configuration of pretraining.}
\label{tab:training config1}
\vspace{-1.em}
\end{table}

\begin{table}[!h]
\centering
\begin{tabular}{l|c}
\toprule
\textbf{Hyperparameter} & \textbf{Value}\\
\midrule
optimizer & AdamW \\ 
learning rate & 5e-5\\
batch size & 1 \\ 
gradient accumulation & 256 \\
clip norm & 2.0 \\
training steps & 7.9k \\
dtype & bfloat16 \\
\bottomrule
\end{tabular}
\caption{Detailed configuration of fine-tuning.}
\label{tab:training config2}
\vspace{-1.em}
\end{table}

\subsection{Prompt Template on Evaluation}
The prompt template we used for evaluation is as follows:

\texttt{Response the Prompt based on the below text:\textbackslash n\textbackslash n \{context\}\textbackslash n\textbackslash n Prompt:\{prompt\}}

\section{Profiling Setup}
We use a single Nvidia RTX A6000 GPU (48GB) for pretraining, fine-tuning, and efficiency tests (Table \ref{tab:overhead}). 
The CPU of our machine is Intel(R) Xeon(R) Gold 6326 with 16 cores and 1007GB RAM. 
The runtime configuration is python=3.8.18, pytorch=1.13.1, cuda=11.7, cudnn=8.5.

\section{Theoretical Analysis}\label{app:analysis}
\subsection{Complexity Analysis}
\begin{figure}[!h]
  \centering
  \includegraphics[width=\linewidth]{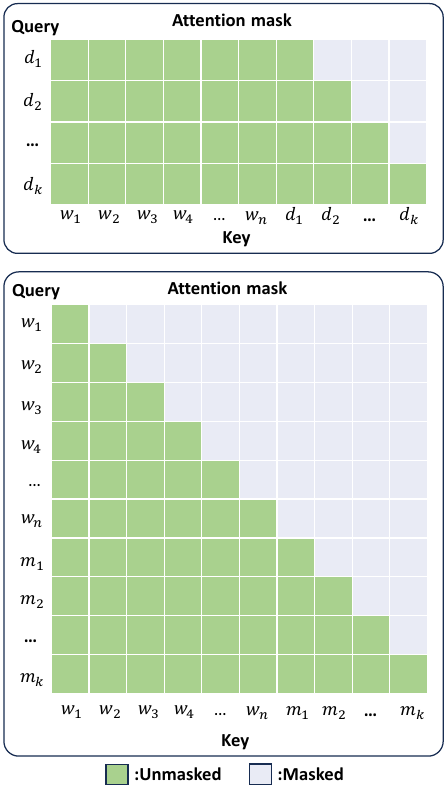}
  \vspace{-1.em}
  \caption{\textbf{Top:} Attention mask in IC-Former. \textbf{Bottom:} Attention mask in ICAE. The $d_{i}$ represents digest tokens in IC-Former and the $m_{i}$ represents the memory tokens in ICAE's encoder.}
  \label{fig:mask}
  \vspace{-1.em}
\end{figure}

In Table \ref{tab:complexity} we assert that the time and space complexity of ICAE is $\mathcal{O}(n^2 + 2kn)$. 
This conclusion can be easily drawn by comparing the attention maps of the IC-Former and ICAE.
As illustrated in Figure \ref{fig:mask}, ICAE utilizes memory tokens and context for causal self-attention interaction, resulting in a complexity of $\mathcal{O}\left((n+k)^2\right)\sim \mathcal{O}(n^2+2kn)$.

\subsection{Floating Point Operations Calculation}
When calculating the floating-point operations, we considered only the matrix multiplication computations involved in the attention and feed-forward network (FFN) modules, while ignoring the relatively smaller computational overhead of modules such as normalization and softmax.

Given context embedding with shape of $[b,s,h]$ where $b$ represents batch size, $s$ represents sequence length and $h$ represents hidden size, the theoretical calculation of the FLOPs for ICAE and IC-Former required to compress it into vectors of length $k$ are shown in Tables \ref{tab:icae_flops} \& \ref{tab:icformer_flops}:

\begin{table}[!h]
\centering
\begin{tabular}{c|c}
\toprule
\textbf{Modules} & \textbf{FLOPs} \\
\midrule
$xW_{Q}/W_{K}/W_{V}$ & $3\cdot 2b(s+k)h^2$ \\
$QK^T$ & $2b(s+k)^2 h$ \\
$AV$ & $2b(s+k)^2 h$ \\
$xW_O$ & $2b(s+k)h^2$ \\
\midrule
$x_{out}W_{up}$ & $2b(s+k)hm$ \\
$x_{out}W_{gate}$ & $2b(s+k)hm$ \\
$x_{out}W_{down}$ & $2b(s+k)hm$ \\
\midrule
\multirow{2}{*}{SUM} & $4bh(s+k)(2h+s+k)$ \\
& $+6bhm(s+k)$ \\
\bottomrule
\end{tabular}
\caption{Theoretical complexity in each layer of ICAE's encoder. $A$ represents the attention scores matrix, $m$ represents the intermediate size of FFN.}
\label{tab:icae_flops}
\vspace{-1.em}
\end{table}
\begin{table}[!h]
\centering
\begin{tabular}{c|c}
\toprule
\textbf{Modules} & \textbf{FLOPs} \\
\midrule
$xW_{Q}$ & $2bkh^2$ \\
$xW_{K}/W_{V}$ & $2\cdot 2b(s+k)h^2$ \\
$QK^T$ & $2bk(s+k) h$ \\
$AV$ & $2bk(s+k) h$ \\
$xW_O$ & $2bkh^2$ \\
\midrule
$x_{out}W_{up}$ & $2bkhm$ \\
$x_{out}W_{gate}$ & $2bkhm$ \\
$x_{out}W_{down}$ & $2bkhm$ \\
\midrule
\multirow{2}{*}{SUM} & $4bkh^2+2bh(s+k)(h+2k)$ \\
& $+6bkhm$ \\
\bottomrule
\end{tabular}
\caption{Theoretical complexity in each layer of IC-Former. $A$ represents the attention scores matrix, $m$ represents the intermediate size of FFN.}
\label{tab:icformer_flops}
\vspace{-1.em}
\end{table}

The ratio of FLOPs between ICAE and IC-Former $R$ can be calculated as follows:
\begin{align}
    R = \frac{l_1\cdot [2(s+k)(2h+s+k)+3m(s+k)]}{l_2\cdot[2kh+(s+k)(h+2k)+3mk]},
\end{align}
where $l_1$ is the layers of ICAE and $l_2$ is the layers of IC-Former.

In our experimental settings, $l_1=32$, $l_2=3$, $s=512$, $k=128$, $h=4096$, $m=11004$, thus
\begin{equation}
    R \approx 32.39
\end{equation}

\section{Case Study}\label{app:case}
In Table \ref{tab:appendix case}, we present several cases to compare the outputs of Llama2-7b-chat based on the 128 digest vectors generated from the pretrained and non-pretrained IC-Former. 
The results indicate that the IC-Former without pre-training has a poor ability to capture contextual information and thus is more prone to hallucinating.

\section{Attention Maps in IC-Former}\label{app:attention map}
\begin{figure*}
    \centering
    \begin{subfigure}{\linewidth}
        \includegraphics[width=\linewidth]{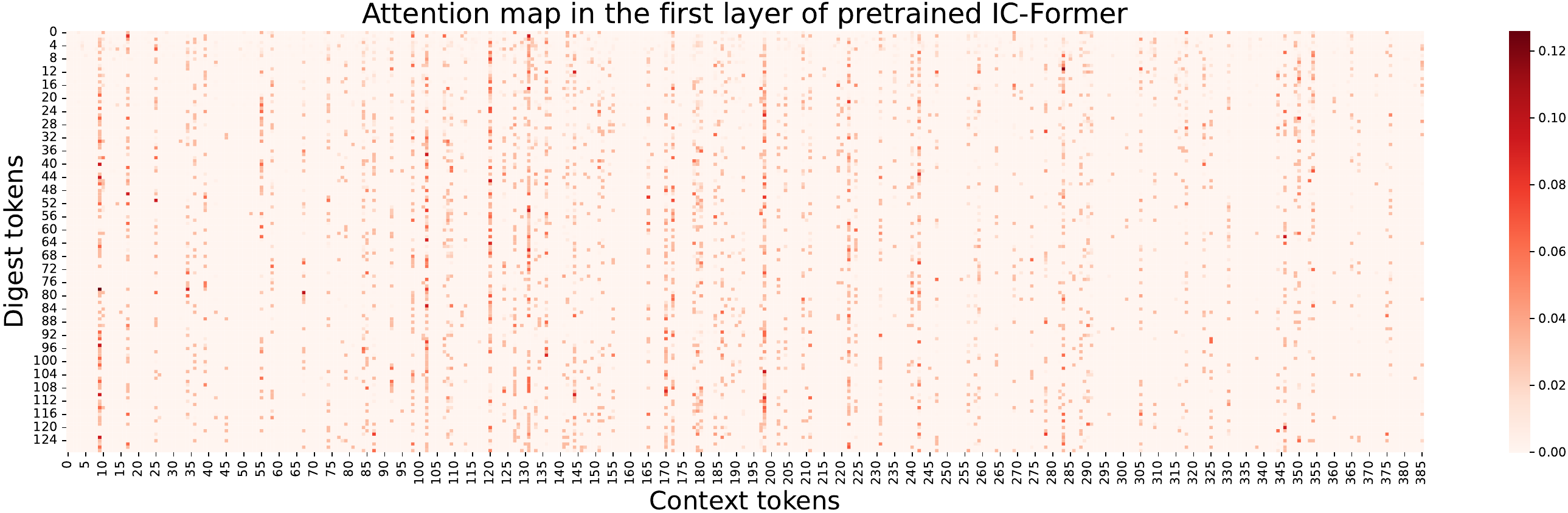}
        \vspace{1em} 
    \end{subfigure}
    \begin{subfigure}{\linewidth}
        \includegraphics[width=\linewidth]{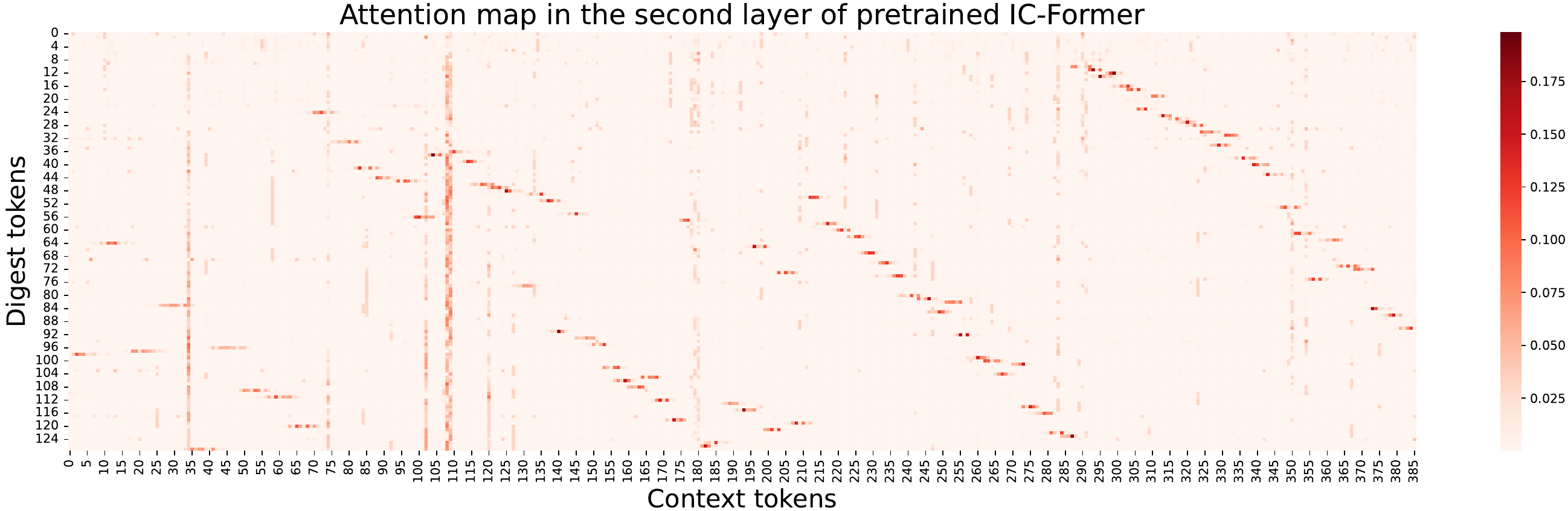}
        \vspace{1em} 
    \end{subfigure}
    \begin{subfigure}{\linewidth}
        \includegraphics[width=\linewidth]{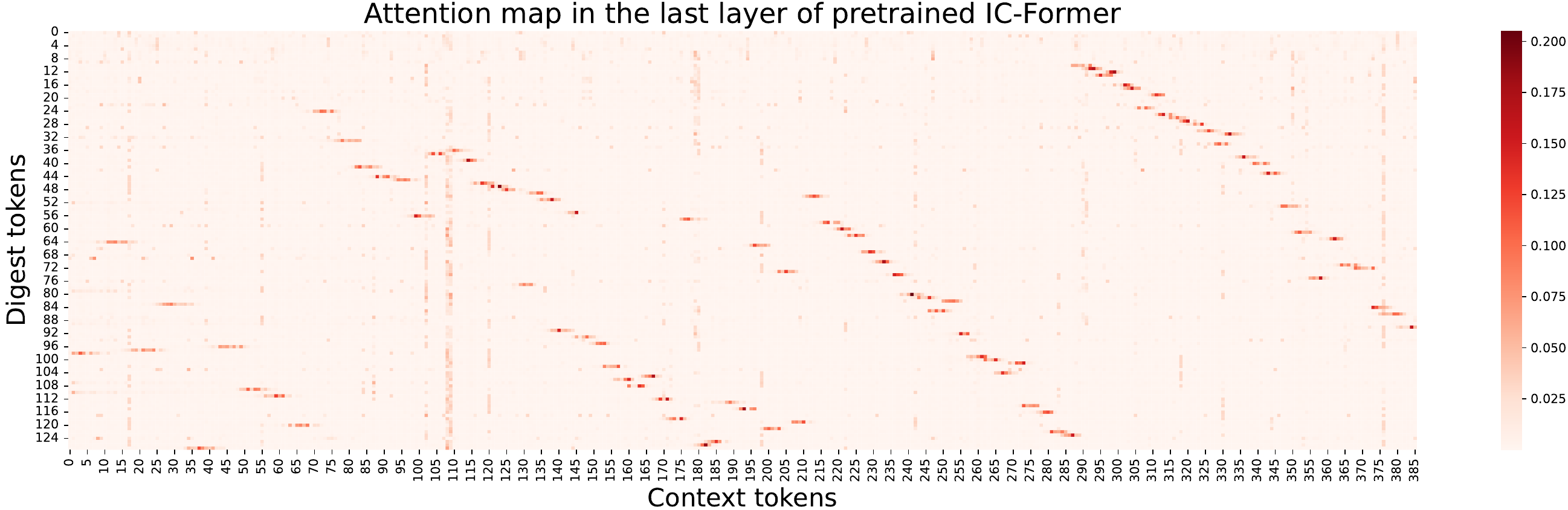}
    \end{subfigure}
    \caption{Complete attention maps of pretrained IC-Former. From top to bottom are attention maps of the first, second, and third layers of IC-Former.}
    \label{fig:pretrain attention}
\end{figure*}

\begin{figure*}
    \centering
    \begin{subfigure}{\linewidth}
        \includegraphics[width=\linewidth]{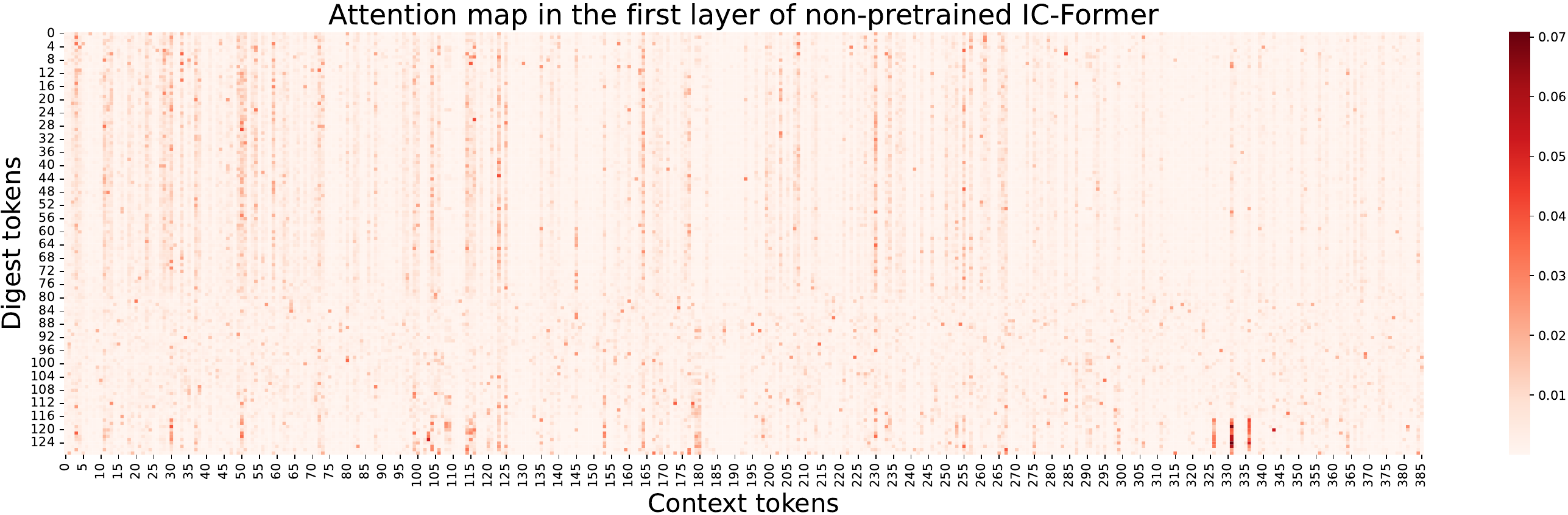}
        \vspace{1em} 
    \end{subfigure}
    \begin{subfigure}{\linewidth}
        \includegraphics[width=\linewidth]{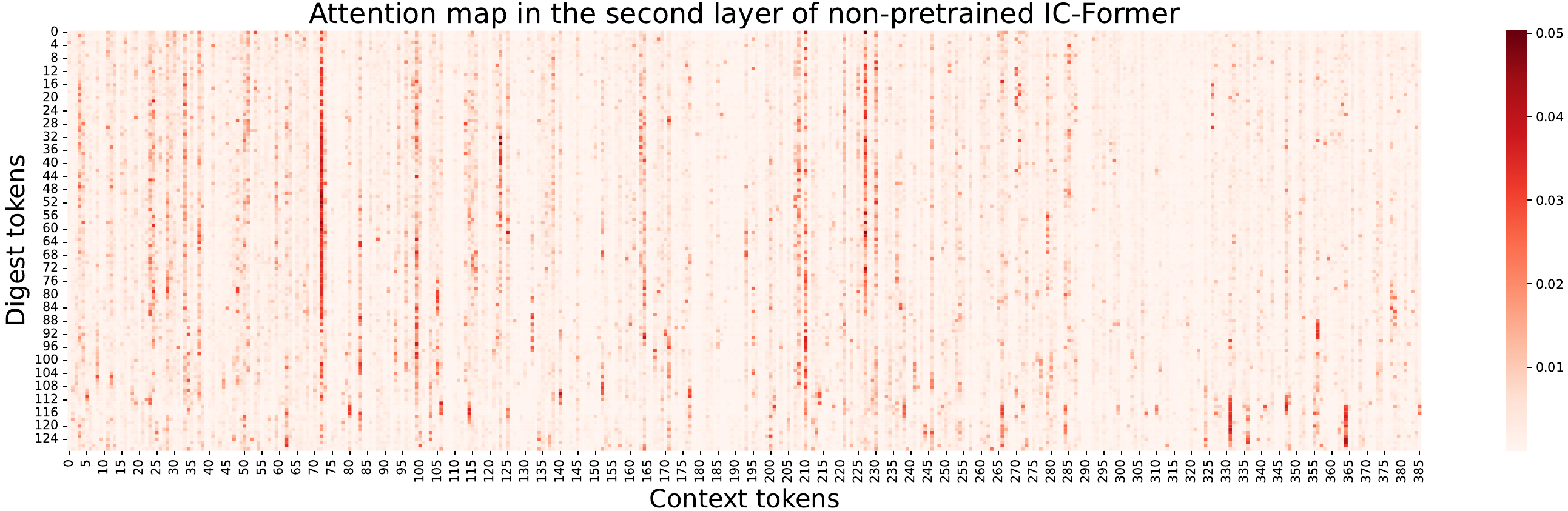}
        \vspace{1em} 
    \end{subfigure}
    \begin{subfigure}{\linewidth}
        \includegraphics[width=\linewidth]{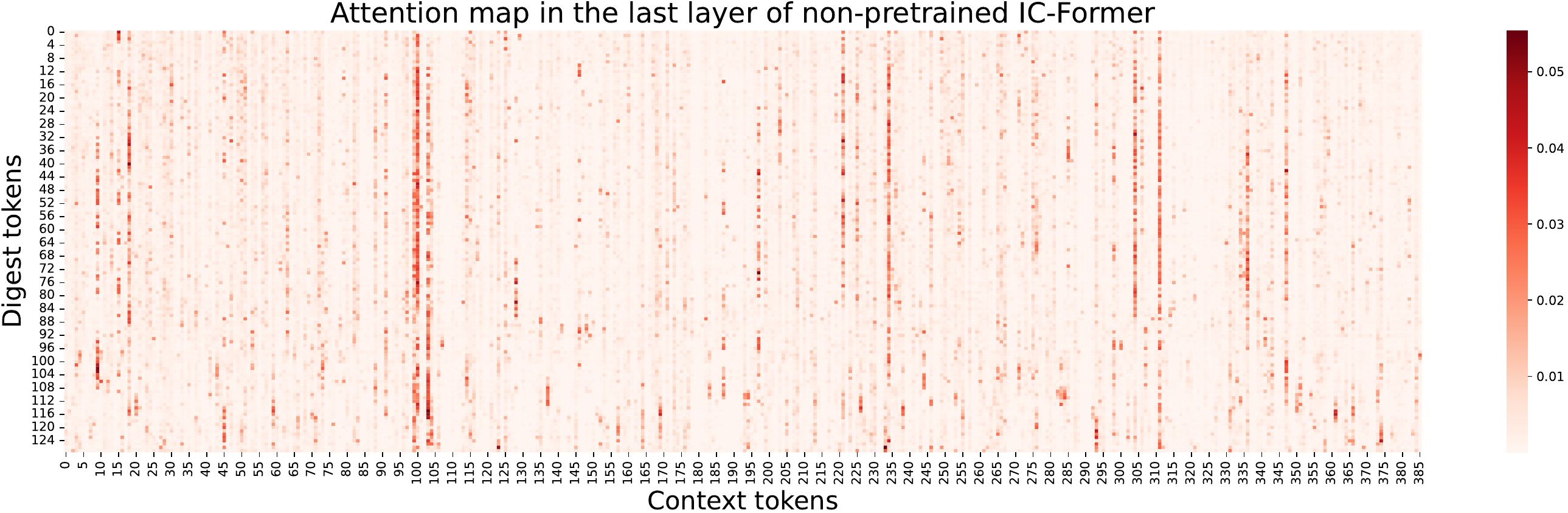}
    \end{subfigure}
    \caption{Complete attention maps of non-pretrained IC-Former. From top to bottom are attention maps of the first, second, and third layers of IC-Former.}
    \label{fig:non-pretrain attention}
\end{figure*}

Additionally, by comparing the attention maps of the pretrained and non-pretrained IC-Former models (Figure \ref{fig:pretrain attention} \& \ref{fig:non-pretrain attention}), it is observable that the non-pretrained IC-Former does not exhibit the phenomenon of neighborhood information aggregation. 
Furthermore, the words captured by each layer do not demonstrate distinct grammatical patterns, which underscores the necessity of pretraining in enhancing model performance.

\begin{table*}
\centering
\begin{tabularx}{2\columnwidth}{>{\arraybackslash}X}
\toprule
\textbf{Context 1} \\
\midrule
French senior civil servant arrested on suspicion of spying for North Korea \\ \\
November 27, 2018 by Joseph Fitsanakis \\ \\
A senior civil servant in the upper house of the French parliament has been arrested on suspicion of spying for North Korea, according to prosecutors. The news of the suspected spy’s arrest was first reported on Monday by Quotidien, a daily politics and culture show on the Monaco-based television channel TMC. The show cited “a judicial source in Paris” and said that France’s domestic security and counterintelligence agency, the General Directorate for Internal Security (DGSI), was in charge of the espionage case. \\
\textbf{(omitted 155 tokens)} \\
\textcolor{blue}{Quennedey is believed to be the president of the Franco-Korean Friendship Association,} the French branch of a Spanish-based organization that lobbies in favor of international support for North Korea. \\
\midrule
\textbf{Prompt:} Mention the organization Benoit Quennedey is believed to be the president of.\\
\midrule
\textbf{Output (with pretrained IC-Former):} Benoit Quennedey is believed to be the president of the Franco-Korean Friendship Association. \\
\midrule
\textbf{Output (with non-pretrianed IC-Former):} Benoit Quennedey is believed to be the president of the \textcolor{red}{French intelligence agency, the Directorate General for External Security (DGSE).} \\
\midrule
\textbf{Label (by GPT-4):} Franco-Korean Friendship Association \\
\toprule
\textbf{Context 2}\\
\midrule
One of NewsBusters' classification categories is "Covert Liberal Activists," and my antennae immediately went up when Lauren Leader was introduced on today's Morning Joe as "co-founder and CEO of All In Together, "a non-partisan, women's civic and political organization." \\
\textbf{(omitted 462 tokens)} \\
Another clue as to Leader's liberal leanings. \textcolor{blue}{Questioning whether the immigrants "actually threw rocks," she cites . . . NPR reporting that things were "relatively peaceful except that, obviously, they were trying to break through."} When people are trying to illegally "break through" a border, there is nothing "relatively peaceful" about it.\\
\midrule
\textbf{Prompt:} What source does Lauren Leader use to question the accuracy of the immigrants throwing rocks? \\
\midrule
\textbf{Output (with pretrained IC-Former):} Lauren Leader uses NPR to question the accuracy of the immigrants throwing rocks. \\
\midrule
\textbf{Output (with non-pretrianed IC-Former):} The source is a \textcolor{red}{Fox News segment.} \\
\midrule
\textbf{Label (by GPT-4):} Lauren Leader cites NPR reporting as a source to question the accuracy of the immigrants throwing rocks. \\
\bottomrule
\end{tabularx}
\caption{Examples of output results from Llama2-7b-chat model utilizing digest vectors generated by pretrained and non-pretrained IC-Former models. The evidence of prompt is marked in \textcolor{blue}{blue} and \textcolor{red}{red} denote the outputs that are not faith to the original context.}
\label{tab:appendix case}
\end{table*}

\end{document}